\documentclass[11pt,a4paper]{article}
\pdfoutput=1
\usepackage{xr-hyper}
\usepackage[hyperref]{acl2020}
\usepackage{times}
\usepackage{latexsym}

\usepackage{microtype}
\usepackage{url}
\usepackage{graphicx}
\usepackage{amssymb}
\usepackage{comment}
\usepackage{booktabs}
\usepackage{amsmath}
\usepackage{enumitem}
\usepackage{caption}
\usepackage{subcaption}
\usepackage{booktabs}
\usepackage{multirow}
  
\usepackage{algorithmicx}
\usepackage{algpseudocode}
\usepackage{algorithm}
\usepackage{placeins}
\usepackage{xcolor, soul}
\sethlcolor{pink}
\usepackage[export]{adjustbox}
\usepackage[nameinlink]{cleveref}

\usepackage{amsmath,amsfonts,bm}

\newcommand{\mor}{\textsc{Morpheus}}

\newcommand{\maxinflex}{\textsc{MaxInflected}}
\newcommand{\mytilde}{{\raise.17ex\hbox{$\scriptstyle\mathtt{\sim}$}}}

\newcommand{\red}[1]{\textcolor{red}{#1}}

\def\eqref#1{equation~\ref{#1}}

\def\1{\bm{1}}

\DeclareMathAlphabet{\mathsfit}{\encodingdefault}{\sfdefault}{m}{sl}
\SetMathAlphabet{\mathsfit}{bold}{\encodingdefault}{\sfdefault}{bx}{n}

\def\gD{{\mathcal{D}}}

\def\gL{{\mathcal{L}}}

\DeclareMathOperator*{\argmax}{arg\,max}

\aclfinalcopy 
\def\aclpaperid{606} 

\title{It's Morphin' Time! \\Combating Linguistic Discrimination with Inflectional Perturbations}

\author{Samson Tan$^\S$$^\natural$, Shafiq Joty$^{\ddagger}$$^\S$, Min-Yen Kan$^\natural$, Richard Socher$^{\S}$\\
  $^\S$Salesforce Research\\
  $^\natural$National University of Singapore \\
  $^\ddagger$Nanyang Technological University \\
  $^\S$\texttt{\{samson.tan,sjoty,rsocher\}@salesforce.com} \\ 
  $^\natural$\texttt{kanmy@comp.nus.edu.sg}
}

\date{}

\begin{document}
\maketitle

\begin{abstract}
Training on only perfect Standard English corpora predisposes pre-trained neural networks to discriminate against minorities from non-standard linguistic backgrounds (e.g., African American Vernacular English, Colloquial Singapore English, etc.). We perturb the inflectional morphology of words to craft plausible and semantically similar adversarial examples that expose these biases in popular NLP models, e.g., BERT and Transformer, and show that adversarially fine-tuning them for a single epoch significantly improves robustness without sacrificing performance on clean data.\footnote{Code and adversarially fine-tuned models available at \scriptsize{ \urlstyle{tt}\url{https://github.com/salesforce/morpheus}}.}
\end{abstract}

\section{Introduction} 
\label{sec:intro}
In recent years, Natural Language Processing (NLP) systems have gotten increasingly better at learning complex patterns in language by pre-training large language models like BERT, GPT-2, and CTRL \cite{devlin2018bert,radford2019language,keskarCTRL2019}, and fine-tuning them on task-specific data to achieve state of the art results has become a norm. However, deep learning models are only as good as the data they are trained on. 

Existing work on societal bias in NLP primarily focuses on attributes like race and gender \citep{bolukbasi2016,may-etal-2019-measuring}. In contrast, we investigate a uniquely NLP attribute that has been largely ignored: linguistic background.

Current NLP models seem to be trained with the implicit assumption that everyone speaks fluent (often U.S.) Standard English, even though two-thirds ($>$700 million) of the English speakers in the world speak it as a second language (L2) \citep{ethno2019}. Even among native speakers, a significant number speak a dialect like African American Vernacular English (AAVE) rather than Standard English \citep{crystal2003english}. In addition, these World Englishes exhibit variation at multiple levels of linguistic analysis \citep{KachruKN2009}.

Therefore, putting these models directly into production without addressing this inherent bias puts them at risk of committing linguistic discrimination by performing poorly for many speech communities (e.g., AAVE and L2 speakers). This could take the form of either failing to understand these speakers \citep{ling-on-trial,tatman-2017-gender}, or misinterpreting them. For example, the recent mistranslation of a minority speaker's social media post resulted in his wrongful arrest \citep{translation-arrest2017}.

\begin{figure}[t]
    \includegraphics[width=0.483\textwidth, right]{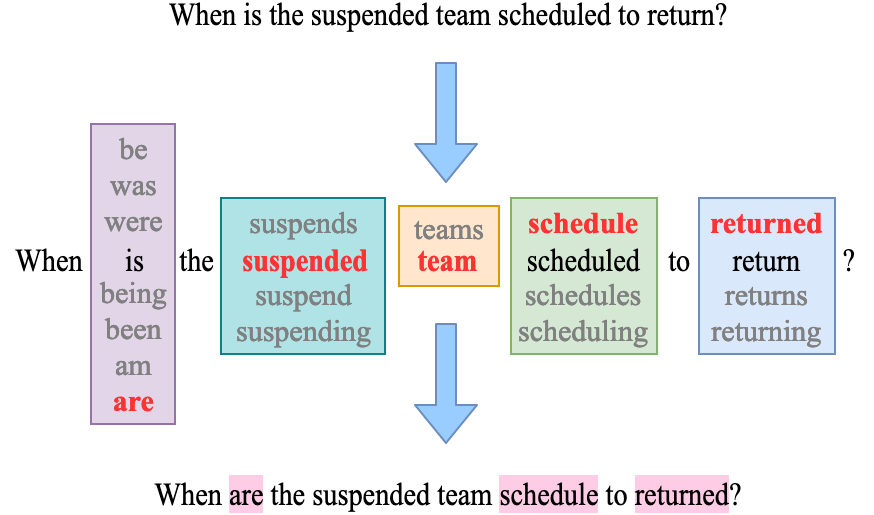}
    \caption{\mor\ looks at each noun, verb, or adjective in the sentence and selects the inflected form (marked in \textbf{\red{red}}) that maximizes the target model's loss. To maximize semantic preservation, \mor\ only considers inflections belonging to the same universal part of speech as the original word.}
    \label{fig:morpheus}
\end{figure}{}

Since L2 (and many L1 dialect) speakers often exhibit variability in their production of inflectional morphology\footnote{\href{https://www.britannica.com/topic/inflection}{Inflections} convey tense, quantity, etc. See \Cref{appendix:dialect} for dialectal examples.} \citep{lardiere1998case,white2000,haznedar2002,white2003,seymourAAVE}, we argue that NLP models should be robust to inflectional perturbations in order to minimize their chances of propagating linguistic discrimination. Hence, in this paper, we:

\begin{itemize}[leftmargin=*]
\vspace{-0.2em}
\itemsep0em
    \item Propose \mor, a method for generating plausible and semantically similar adversaries by perturbing the inflections in the clean examples (\Cref{fig:morpheus}). In contrast to recent work on adversarial examples in NLP \citep{belinkov2018synthetic,ebrahimi-etal-2018-hotflip,SinghGR18},
    we exploit morphology to craft our adversaries.
    \item Demonstrate its effectiveness on multiple machine comprehension and  translation models, including BERT and Transformer (Tables \ref{tab:examples} \& \ref{tab:morpheus_results}).
    \item Show that adversarially fine-tuning the model on an adversarial training set generated via weighted random sampling is sufficient for it to acquire significant robustness, while preserving performance on clean examples (\Cref{tab:adv_training_results}).
\end{itemize}
\vspace{-0.2em}
To the best of our knowledge, we are the first to investigate the robustness of NLP models to inflectional perturbations and its ethical implications.

\section{Related Work} \label{sec:bgr}

\paragraph{Fairness in NLP.} 
It is crucial that NLP systems do not amplify and entrench social biases \citep{hovy-spruit-2016-social}. Recent research on fairness has primarily focused on racial and gender biases within distributed word representations \citep{bolukbasi2016}, coreference resolution \citep{gender-bias-in-coreference-resolution2018}, sentence encoders \citep{may-etal-2019-measuring}, and language models \citep{bordia-bowman-2019-identifying}. However, we posit that there exists a significant potential for \emph{linguistic} bias that has yet to be investigated, which is the motivation for our work.

\paragraph{Adversarial attacks in NLP.} 
First discovered in computer vision by \citet{szegedy2014}, adversarial examples are data points crafted with the intent of causing a model to output a wrong prediction. In NLP, this could take place at the character, morphological, lexical, syntactic, or semantic level.

\citet{jia-liang-2017-adversarial} showed that question answering models could be misled into choosing a distractor sentence in the passage that was created by replacing key entities in the correct answer sentence. \citet{belinkov2018synthetic} followed by demonstrating the brittleness of neural machine translation systems against character-level perturbations like randomly swapping/replacing characters. However, these attacks are not optimized on the target models, unlike \citet{ebrahimi-etal-2018-hotflip}, which makes use of the target model's gradient to find the character change that maximizes the model's error. 

Since these attacks tend to disrupt the sentence's semantics, \citet{SinghGR18} and \citet{michel-etal-2019-evaluation} propose searching for adversaries that preserve semantic content. \citet{alzantot-etal-2018-generating} and \citet{jin2019bert} explore the use of synonym substitution to create adversarial examples, using word embeddings to find the $n$ nearest words. \citet{Eger2019TextPL} take a different approach, arguing that adding visual noise to characters leaves their semantic content undisturbed. \citet{iyyer-etal-2018-adversarial} propose to create paraphrase adversaries by conditioning their generation on a syntactic template, while \citet{ZhangBH19} swap key entities in the sentences. \citet{zhang2019adversarial} provide a comprehensive survey of this topic.

\vspace{-0.1em}
\paragraph{Adversarial training.} In order to ensure our NLP systems are not left vulnerable to powerful attacks, most existing work make use of adversarial training to improve the model's robustness \citep{goodfellow2015}. This involves augmenting the training data either by adding the adversaries to or replacing the clean examples in the training set.

\vspace{-0.1em}
\paragraph{Summary.} Existing work in fairness mostly focus on tackling bias against protected attributes like race and gender, while those in adversarial NLP primarily investigate character- and word-level perturbations and seek to improve the models' robustness by retraining them from scratch on the adversarial training set. Our work makes use of perturbations in \emph{inflectional morphology} to highlight the linguistic bias present in models such as BERT and Transformer, before showing that simply \emph{fine-tuning} the models for \emph{one} epoch on the adversarial training set is sufficient to achieve significant robustness while maintaining performance on clean data.

\section{Generating Inflectional Perturbations} \label{sec:method}

\begin{table*}[t!]
\small
    \centering
    \begin{tabular}{r  p{0.85\textwidth}}
    \toprule
    \multicolumn{2}{c}{\textbf{Extractive Question Answering}} \\
    \midrule
    Original & When is the suspended team scheduled to return? \\
    Adversary & When \hl{are} the suspended team \hl{schedule} to \hl{returned}? \\
    Prediction & \textbf{Before:} 2018 \quad \textbf{After:} No answer \\
    \midrule
        Original & Who upon arriving gave the original viking settlers a common identity?  \\
    Adversary & Who upon \hl{arrive} \hl{give} the original viking \hl{settler} a common \hl{identities}? \\
    Prediction & \textbf{Before:} Rollo \quad \textbf{After:} almost no foreign settlers \\
    \midrule
    \multicolumn{2}{c}{\textbf{Neural Machine Translation}} \\
    \midrule
    Original & Israeli warplanes struck a target inside the Syrian port city of Latakia Thursday night, a senior administration official confirms to Fox News. \\
    Adversary & Israeli warplanes \hl{strikes} a target inside the Syrian port city of Latakia Thursday night, a senior administration official confirms to \hl{Foxes} News. \\
    \multirow[t]{2}{*}{Prediction} & \textbf{Before:} Un haut responsable de l'administration confirme à Fox News que des avions de combat israéliens ont frappé une cible à l'intérieur de la ville portuaire syrienne de Lattaquié dans la nuit de jeudi. \\
    & \textbf{After:} Le président de la République, Nicolas Sarkozy, a annoncé jeudi que le président de la République, Nicolas Sarkozy, s'est rendu en République démocratique du Congo. \\
    \bottomrule
    \end{tabular}
    \vspace{-0.5em}
    \caption{Adversarial examples found for BERT, SpanBERT, and Transformer-big. While not perfectly grammatical, it is plausible for English dialect and second language (L2) speakers to produce such sentences.\\ \textbf{(Top)} Models trained on SQuAD 2.0 are more fragile than those trained on SQuAD 1.1, 
    and have a bias towards predicting ``no answer". 
    Examples are answerable questions and therefore present in both SQuAD 1.1 and 2.0.\\\textbf{(Bottom)} Perturbing two inflections caused Transformer-big to output a completely irrelevant sentence. In addition, adversarial examples for \mytilde1.4\% of the test set caused the model to output the source (English) sentences.}
    \label{tab:examples}
\end{table*}

\vspace{-0.1em}
Inflectional perturbations inherently preserve the general semantics of a word since the root remains unchanged. In cases where a word's part of speech (POS) is context-dependent (e.g., \emph{duck} as a verb or a noun), restricting perturbations to the original POS further preserves its original meaning.

Additionally, since second language speakers are prone to inflectional errors \citep{haznedar2002,white2003}, adversarial examples that perturb the inflectional morphology of a sentence should be \emph{less perceivable} to people who interact heavily with non-native speakers or are themselves non-native speakers. Hence, we present \mor, our proposed method for crafting inflectional adversaries.

\subsection{\mor: A Greedy Approach}\label{sec:morpheus}
\paragraph{Problem formulation.} Given a target model $f$ and an original input example $x$ for which the ground truth label is $y$, our goal is to generate the adversarial example $x'$ that maximizes $f$'s loss. Formally, we aim to solve the following problem:
\begin{equation}
x' = \underset{x_c}{\argmax} \; \gL(y,f(x_c))  \label{eq:loss} 
\end{equation}
where $x_c$ is an adversarial example generated by perturbing $x$, $f(x)$ is the model's prediction, and $\gL(\cdot)$ is the model's loss function. In this setting, $f$ is a neural model for solving a specific NLP task.

\paragraph{Proposed solution.} 
To solve this problem, we propose \mor\ (\Cref{alg:morpheus}), an approach that greedily searches for the inflectional form of each \emph{noun,} \emph{verb,} or \emph{adjective} in $x$ that maximally increases $f$'s loss (Eq. \ref{eq:loss}). For each token in $x$, \mor\ calls \maxinflex\ to find the inflected form that caused the greatest increase in $f$'s loss.\footnote{{A task-specific evaluation metric may be used instead of the loss in situations where it is unavailable.} However, as we discuss later, the choice of metric is important for optimal performance and should be chosen wisely.} \Cref{tab:examples} presents some adversarial examples obtained by running \mor\ on state-of-the-art machine reading comprehension and translation models: namely, BERT \cite{devlin2018bert}, SpanBERT \citep{spanbert19}, and Transformer-big \citep{vaswani2017attention,ott2018-scaling}.

\begin{algorithm}[t]
\small
\begin{algorithmic}
\Require Original instance $x$, Label $y$, Model $f$
\Ensure Adversarial example $x'$

\State $T \gets \Call{tokenize}{x}$
\ForAll {$i = 1, \ldots, |T|$}
    \If{$\Call{POS}{T_i}\in\! \{\text{\small NOUN},\text{\small VERB},\text{\small ADJ}\}$}
        \State $I \gets \Call{GetInflections}{T_i}$
        \State $T_i \gets \Call{MaxInflected}{I,T,y,f}$
    \EndIf
\EndFor
\State $x' \gets \Call{detokenize}{T}$
\State \Return $x'$
\caption{\mor}
\label{alg:morpheus}
\end{algorithmic}
\end{algorithm}

There are two possible approaches to implementing \textsc{MaxInflected}: one is to modify each token independently from the others in \emph{parallel}, and the other is to do it \emph{sequentially} such that the increase in loss is accumulated as we iterate over the tokens. 
A major advantage of the parallel approach is that it is theoretically possible to speed it up by ${t}$ times, where ${t}$ is the number of tokens which are nouns, verbs, or adjectives. However, since current state-of-the-art models rely heavily on contextual representations, the sequential approach is likely to be more effective in finding combinations of inflectional perturbations that cause major increases in loss. We found this to be the case in our preliminary experiments (see \Cref{tab:parallel_seq} in \Cref{appendix:misc_figs}).

\paragraph{Assumptions.} \mor\ treats the target model as a \emph{black box} and maximally requires only access to the model's logits to compute the loss. As mentioned, task-specific metrics may be used instead of the loss as long as the surface is not overly ``flat'', like in a step function. Examples of inappropriate metrics are the exact match and F$_1$ scores for extractive question answering, which tend to be 1 for most candidates but drop drastically for specific ones. This may affect \mor' ability to find an adversary that induces absolute model failure.

While the black box assumption has the advantage {of not requiring access to the target model's gradients and parameters},  
a limitation is that we need to query the model for each candidate inflection's impact on the loss, as opposed to \citet{ebrahimi-etal-2018-hotflip}'s approach. However, this is not an issue for inflectional perturbations since each word \emph{usually} has less than 5 possible inflections.

\paragraph{Candidate generation.} We make use of {\tt lemminflect}\footnote{\scriptsize{ \urlstyle{tt}\url{https://github.com/bjascob/LemmInflect}}} to generate candidate inflectional forms in the \textsc{GetInflections} method, a simple process in which the token is first lemmatized before being inflected. In our implementation of \textsc{GetInflections}, we also allow the user to specify if the candidates should be constrained to the same universal part of speech. 

\paragraph{Semantic preservation.} \textsc{Morpheus} constrains its search to inflections belonging to the same \emph{universal} part of speech. For example, take the word ``duck". Depending on the context, it may either be a verb or a noun. In the context of the sentence ``There's a jumping duck", ``duck" is a noun and\textsc{Morpheus} may only choose alternate inflections associated with nouns. 

This has a higher probability of preserving the sentence's semantics compared to most other approaches, like character/word shuffling or synonym swapping, since the root word and its position in the sentence remains unchanged. 

\paragraph{Early termination.} \mor\ selects an inflection if it increases the loss.  In order to avoid unnecessary searching, it terminates once it finds an adversarial example that induces model failure. In our case, we define this as a score of 0 on the task's evaluation metric (the higher, the better).

\paragraph{Other implementation details.} In order to increase overall inflectional variation in the set of adversarial examples, \textsc{GetInflections} shuffles the generated list of inflections before returning it (see \Cref{fig:shuffling} in Appendix). Doing this has no effect on \textsc{Morpheus}' ability to induce misclassification, but prevents overfitting during adversarial fine-tuning, which we discuss later in \Cref{sec:adv-train}. Additionally, since \mor\ greedily perturbs each eligible token in $x$, it may get stuck in a local maximum for some $x$ values. To mitigate this, we run it again on the reversed version of $x$ if the early termination criterion was not fulfilled during the forward pass. 

Finally, we use {\tt sacremoses}\footnote{\scriptsize{\urlstyle{tt}\url{https://github.com/alvations/sacremoses}}} for tokenization and NLTK \citep{nltk09} for POS tagging.

\section{Experiments}\label{sec:experiments}

\begin{table*}[t!]
\small
    \centering
    \begin{tabular}{l l c c c}
         \toprule
         Dataset & Model & Clean & Random & \mor\ \\
         \midrule
         \multirow{6}{0.3\textwidth}{SQuAD 2.0 Answerable Questions (F$_1$)} & GloVe-BiDAF & 78.67 & 74.00 ($-$5.93\%) & \textbf{53.94 ($-$31.43\%)} \\
         & ELMo-BiDAF & 80.90 & 76.81 ($-$5.05\%) & 62.17 ($-$23.15\%)\\
         & BERT$_{\text{SQuAD 1.1}}$ & 93.14 & 90.90 ($-$2.40\%) & 82.79 ($-$11.11\%) \\
         & SpanBERT$_{\text{SQuAD 1.1}}$ & 91.88 & 91.61 ($-$0.29\%) & 82.86 ($-$9.81\%) \\
         & BERT$_{\text{SQuAD 2}}$ & 81.19 & 74.13 ($-$8.69\%) & \textbf{57.47 ($-$29.21\%)} \\
         & SpanBERT$_{\text{SQuAD 2}}$ & 88.52 & 84.88 ($-$4.11\%) & 69.47 ($-$21.52\%) \\
         \midrule
         \multirow{2}{0.23\textwidth}{SQuAD 2.0 All Questions (F$_1$)} & BERT$_{\text{SQuAD 2}}$ & 81.52 & 78.87 ($-$3.25\%)  & \textbf{67.24 ($-$17.51\%)} \\
         & SpanBERT$_{\text{SQuAD 2}}$ & 87.71 & 85.46 ($-$2.56\%) & 73.26 ($-$16.47\%) \\
         \midrule
         \multirow{2}{0.2\textwidth}{newstest2014 En-Fr (BLEU)} & ConvS2S & 40.83 & 27.72 ($-$32.10\%) & \textbf{17.31 ($-$57.60\%)}\\
         & Transformer-big & 43.16 & 30.41 ($-$29.54\%) & 20.57 ($-$56.25\%)\\
         \bottomrule
    \end{tabular}
    \caption{Results for \mor\ on QA and NMT models. The subscript in Model$_{\text{dataset}}$ indicates the dataset used to fine-tune the model. Negated \% decrease w.r.t. the scores on clean data are reported in parentheses for easy comparison across models. Bolded values indicate the largest \% decrease.}
    \label{tab:morpheus_results}
\end{table*}

\paragraph{NLP tasks.} To evaluate the effectiveness of \mor\ at inducing model failure in NLP models, we test it on two popular NLP tasks: question answering (QA) and machine translation (MT). QA involves language understanding (classification), while MT also involves language generation. Both are widely used by consumers of diverse linguistic backgrounds and hence have a high chance of propagating discrimination.

\paragraph{Baseline.} In the below experiments, we include a \textit{random baseline} that randomly inflects each eligible word in each original example.

\paragraph{Measures.} In addition to the raw scores, we also report the \emph{relative decrease} for easier comparison across models since they perform differently on the clean dataset. Relative decrease ($d_r$) is calculated using the following formula: 

\begin{equation}
d_r = \frac{\text{score}_{\text{original}} - \text{score}_{\text{adversarial}}}{\text{score}_{\text{original}}}    
\end{equation}

\subsection{Extractive Question Answering}
Given a question and a passage containing spans corresponding to the correct answer, the model is expected to predict the span corresponding to the answer. Performance for this task is computed using \emph{exact match} or \emph{average F$_1$} \citep{rajpurkar-etal-2016-squad}. We evaluate the effectiveness of our attack using average F$_1$, which is more forgiving (for the target model). From our experiments, the exact match score is usually between 3-9 points lower than the average F$_1$ score.

\paragraph{SQuAD 1.1 and 2.0.} The Stanford Question Answering Dataset (SQuAD) comprises over 100,000 question--answer pairs written by crowdworkers based on Wikipedia articles. SQuAD 1.1 guarantees that the passages contain valid answers to the questions posed \citep{rajpurkar-etal-2016-squad}. SQuAD 2.0 increases the task's difficulty by including another 50,000 unanswerable questions, and models are expected to identify when a passage does not contain an answer for the given question \citep{rajpurkar-etal-2018-know}. Since the test set is not public, we generate adversarial examples from and evaluate the models on the standard dev set.

In addition, the answerable questions from SQuAD 2.0 are used in place of SQuAD 1.1 to evaluate models trained on SQuAD 1.1. This allows for easy comparison between the performance of the SQuAD 1.1-fine-tuned models and SQuAD 2.0-fine-tuned ones for answerable questions. We found performance on the answerable questions from SQuAD 2.0 to be comparable to SQuAD 1.1. 

\paragraph{Models.} We evaluate \mor\ on \citet{Gardner2017AllenNLP}'s implementation of BiDAF \citep{bidaf17}, a common baseline model for SQuAD 1.1, ELMo-BiDAF \citep{elmo18}, the \texttt{transformers} implementation \citep{Wolf2019HuggingFacesTS} of BERT, and SpanBERT, a pre-training method focusing on span prediction that outperforms BERT on multiple extractive QA datasets.

\subsection{Results and Discussion}
From \Cref{tab:morpheus_results}, we see that models based on contextual embeddings (e.g., ELMo and BERT variants) tend to be more robust than those using fixed word embeddings (GloVe-BiDAF). This difference is likely due to the pre-training process, which gives them greater exposure to a wider variety of contexts in which different inflections occur. Removing the POS constraint further degrades the models' performance by another 10\% of the original score, however, this difference is likely due to changes in the semantics and expected output of the examples.
 
\paragraph{BiDAF vs. BERT.} Even after accounting for the performance difference on clean data, the BiDAF variants are significantly less robust to inflectional adversaries compared to the BERT variants. This is likely a result of BERT's greater representational power and masked language modeling pre-training procedure. Randomly masking out words during  pre-training could have improved the models' robustness to small, local perturbations (like ours).

\paragraph{BERT vs. SpanBERT.} In the context of question answering, SpanBERT appears to be slightly more robust than vanilla BERT when comparing overall performance on the two SQuAD datasets. However, the difference becomes significant if we look only at the SQuAD 2.0-fine-tuned models' performance on answerable questions (7\% difference). This indicates that BERT has a stronger bias towards predicting ``no answer'' when it encounters inflectional perturbations compared to SpanBERT. 

\paragraph{SQuAD 1.1 vs. SQuAD 2.0.} The ability to ``know what you don't know" \citep{rajpurkar-etal-2018-know} appears to have been obtained at a great cost. The SQuAD 2.0-fine-tuned models are not only generally less robust to inflectional errors than their SQuAD 1.1 equivalents (6.5\% difference), but also significantly less adept at handling answerable questions (12--18\% difference). This discrepancy suggests a stronger bias in SQuAD 2.0 models towards predicting ``no answer" upon receiving sentences containing inflectional errors (see \Cref{tab:examples}).

As we alluded to earlier, this is particularly troubling: since SQuAD 2.0 presents a more realistic scenario than SQuAD 1.1, it is fair to conclude that such models will inadvertently discriminate against L2 speakers if put into production as is.

\begin{table}[t!]
    \small
    \centering
    \scalebox{0.86}{\begin{tabular}{l l c c}
         \toprule
         \multicolumn{4}{c}{\textbf{SQuAD 2.0 Answerable Questions (F$_1$)}} \\
         \midrule
         Original & Transfer & Clean & \mor\ \\
         \midrule
         \multirow{4}{0.1\textwidth}{GloVe-BiDAF} & BERT$_{\text{SQuAD 1.1}}$ & 93.14 & 89.67 \\
         & SpanBERT$_{\text{SQuAD 1.1}}$ & 91.88 & 90.75 \\
         & BERT$_{\text{SQuAD 2}}$ & 81.19 & 72.21\\
         & SpanBERT$_{\text{SQuAD 2}}$ & 88.52 & 81.95 \\
         \midrule
         \multirow{4}{0.135\textwidth}{BERT$_{\text{SQuAD 1.1}}$} & GloVe-BiDAF & 78.67 & 71.33 \\
         & SpanBERT$_{\text{SQuAD 1.1}}$ & 91.88 & 88.68 \\
         & BERT$_{\text{SQuAD 2}}$ & 81.19 & 69.68 \\
         & SpanBERT$_{\text{SQuAD 2}}$ & 88.52 & 80.11 \\
         \midrule
         \multirow{4}{0.135\textwidth}{SpanBERT$_{\text{SQuAD 1.1}}$} & GloVe-BiDAF & 78.67 & 71.41 \\
         & BERT$_{\text{SQuAD 1.1}}$ & 93.14 & 87.48 \\
         & BERT$_{\text{SQuAD 2}}$ & 81.19 & 70.05 \\
         & SpanBERT$_{\text{SQuAD 2}}$ & 88.52 & 77.89 \\
         \midrule
         \multicolumn{4}{c}{\textbf{SQuAD 2.0 All Questions (F$_1$)}} \\
         \midrule
         Original & Transfer & Clean & \mor\ \\
         \midrule
         BERT$_{\text{SQuAD 2}}$ & SpanBERT$_{\text{SQuAD 2}}$ & 87.71 & 82.49 \\
         SpanBERT$_{\text{SQuAD 2}}$  & BERT$_{\text{SQuAD 2}}$ & 81.52 & 75.54 \\
         
         \bottomrule
    \end{tabular}}
    \caption{Transferability of our adversarial examples.}
    \label{tab:transfer_results}
\end{table}
\paragraph{Transferability.}
Next, we investigate the transferability of adversarial examples found by \mor\ across different QA models and present some notable results in \Cref{tab:transfer_results}. The adversarial examples found for GloVe-BiDAF transfer to a limited extent to other models trained on SQuAD 1.1, however, they have a much greater impact on BERT$_{\text{SQuAD 2}}$ and SpanBERT$_{\text{SQuAD 2}}$ (3--4x more).

We observe a similar pattern for adversarial examples found for SpanBERT$_{\text{SQuAD 1.1}}$. Of the two, BERT is more brittle in general: the SpanBERT$_{\text{SQuAD 1.1}}$ adversaries have a greater effect on BERT$_{\text{SQuAD 2}}$'s performance on answerable questions than on SpanBERT$_{\text{SQuAD 2}}$'s.

{\paragraph{Discussion.} One possible explanation for the SQuAD 2.0 models' increased fragility is the difference in the tasks they were trained for: SQuAD 1.1 models expect all questions to be answerable and only need to contend with finding the right span, while SQuAD 2.0 models have the added burden of predicting whether a question is answerable.} 

Therefore, in SQuAD 1.1 models, the feature space corresponding to a possible answer ends where the space corresponding to another possible answer begins, and there is room to accommodate slight variations in the input (i.e., larger individual spaces). We believe that in SQuAD 2.0 models, the need to accommodate the \texttt{unanswerable} prediction forces the spaces corresponding to the possible answers to shrink, with \texttt{unanswerable} spaces potentially filling the gaps between them. For SQuAD 2.0 models, this increases the probability of an adversarial example ``landing" in the space corresponding to the \texttt{unanswerable} prediction. This would explain the effectiveness of adversarial fine-tuning in \Cref{sec:adv-train}, which intuitively creates a ``buffer" zone and expands the decision boundaries around each clean example.

{The diminished effectiveness of the transferred adversaries at inducing model failure is likely due to each model learning slightly different segmentations of the answer space. As a result, different small, local perturbations have different effects on each model. We leave the in-depth investigation of the above phenomena to future work.}

\subsection{Machine Translation}
We now demonstrate \mor' ability to craft adversaries for NMT models as well, this time \emph{without} access to the models' logits. The WMT'14 English-French test set (newstest2014), containing 3,003 sentence pairs, is used for both evaluation and generating adversarial examples. We evaluate our attack on the {\tt fairseq} implementation of both the Convolutional Seq2Seq \citep{gehring2017convolutional} and Transformer-big models, and report the BLEU score \citep{papineni-etal-2002-bleu} using {\tt fairseq}'s implementation \citep{ott2019fairseq}.

From our experiments (\Cref{tab:morpheus_results}), ConvS2S and Transformer-big appear to be extremely brittle even to inflectional perturbations constrained to the same part of speech (56--57\% decrease). In addition, some adversarial examples caused the models to regenerate the input verbatim instead of a translation: 1.4\% of the test set for Transformer-big, 3\% for ConvS2S (see \Cref{tab:eng_output} in the Appendix for some examples). This is likely due to the joint source/target byte--pair encoding \cite{sennrich2015neural} used by both NMT systems to tackle rare word translation. 

We experimented with both BLEU and chrF \citep{popovic-2015-chrf} as our optimizing criterion\footnote{We use the {\tt sacrebleu} implementation \citep{sacredbleu_post-2018-call}.} and achieved comparable results for both, however, \mor\ found more adversarial examples that caused the model to output random sentences about Nicolas Sarkozy when optimizing for chrF.
\section{Human Evaluation}

\begin{table}
\addtolength{\tabcolsep}{-3pt}
\small
    \centering
    \scalebox{0.75}{\begin{tabular}{l  c c|  c c}
        \toprule
        & \multicolumn{3}{c}{\textbf{Plausibility}} \\
        \midrule
        & \multicolumn{2}{c|}{\textbf{Native U.S. English Speakers}} & \multicolumn{2}{c}{\textbf{Unrestricted}} \\
        & {SQuAD 2.0} & {newstest2014} & {SQuAD 2.0} & {newstest2014} \\
         Native & 11.58\% & 25.64\% & 22.82\% & 32.56\% \\
         L2 Speaker & \textbf{42.82\%} & \textbf{42.30\%} & \textbf{53.58\%} & \textbf{52.82\%} \\
         Beginner & 31.79\% & 23.33\% & 17.17\% & 10.25\% \\
         Non-human & 13.84\% & 8.71\% & 6.41\% & 4.35\% \\
         \midrule
         & \multicolumn{3}{c}{\textbf{Semantic Equivalence}} \\
        \midrule
        & \multicolumn{2}{c|}{\textbf{Native U.S. English Speakers}} & \multicolumn{2}{c}{\textbf{Unrestricted}} \\
        & {SQuAD 2.0} & {newstest2014} & {SQuAD 2.0} & {newstest2014} \\
         Highly Likely & \textbf{52.82\%} & \textbf{62.30\%} & 33.84\% & \textbf{40.76\%} \\
         Likely & 20.51\% & 18.71\% & \textbf{36.15\%} & 33.84\% \\
         Somewhat Likely & 11.02\% & 7.94\% & 22.82\% & 19.48\% \\
         Somewhat Unlikely & 6.92\% & 6.15\% & 5.38\% & 4.35\% \\
         Unlikely & 3.58\% & 3.07\% & 1.53\% & 1.28\% \\
         Highly Unlikely & 5.12\% & 1.79\% & 0.25\% & 0.25\% \\
         \bottomrule
    \end{tabular}}
    \caption{Human judgements for adversarial examples that caused a significant degradation in performance.}
    \label{tab:human_jdgements}
\end{table}

To test our hypothesis that inflectional perturbations are likely to be relatively natural and semantics preserving, we randomly sample 130 adversarial examples\footnote{Only adversarial examples that degraded the F$_1$ score by $>\!50$ and the BLEU score by $>\!15$ were considered.} from each dataset and ask 3 Amazon Mechanical Turk workers to indicate (1) whether the sentences could have been written by a native speaker, L2 speaker, beginner learner\footnote{We define a beginner as one who has just started learning the language, and an L2 speaker to be an experienced speaker.}, or no human; and (2) the likelihood of the original and adversarial examples sharing the same meaning. To ensure the quality of our results, only Turkers who completed $>$10,000 HITs with a $\geq$99\% acceptance rate could access our task. For comparison, we also report ratings by native U.S. English speakers, who were selected via a demographic survey and fluency test adapted from \citet{fluencytest}. Workers were paid a rate of at least \$12/hr.\footnote{Each task was estimated to take 20-25s to be comfortably completed, but they were routinely completed in under 20s.}

\Cref{tab:human_jdgements} shows that Turkers from our unrestricted sample judged \mytilde95\% of our adversaries to be plausibly written by a human and 92\% generally likely to be semantically equivalent to the original examples 92\% of the time, hence validating our hypothesis. Qualitative analysis revealed that ``is/are"$\rightarrow$``am/been" changes accounted for 48\% of the implausible adversaries.

\paragraph{Discussion.} We believe that non-native speakers may tend to rate sentences as more human-like for the following reasons:
\begin{itemize}[leftmargin=*]
\vspace{-0.5em}
\itemsep 0em
    \item Their exposure to another language as a native speaker leads them to accept sentences that mimic errors made by L2 English speakers who share their first language.
    \item Their exposure to the existence of these above-mentioned errors may lead them to be more forgiving of other inflectional errors that are uncommon to them; they may deem these errors as plausibly made by an L2 speaker who speaks a different first language from them.
    \item They do not presume mastery of English, and hence may choose to give the higher score when deciding between 2 choices.
\end{itemize}{}

\begin{figure}[t]
    \centering
    \begin{subfigure}{0.495\textwidth}
        \includegraphics[width=\textwidth]{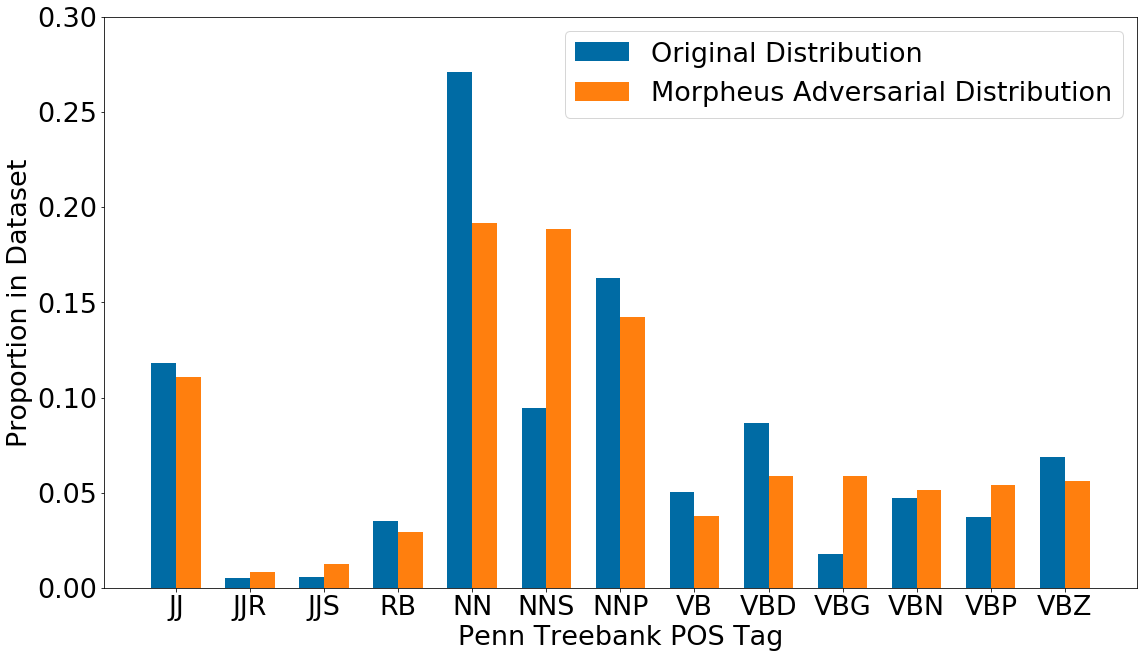}
        \caption{SQuAD 2.0 dev set}
        \label{fig:qa_constr_unconstr}
    \end{subfigure}
    \vfill
     \begin{subfigure}{0.495\textwidth}
        \includegraphics[width=\textwidth]{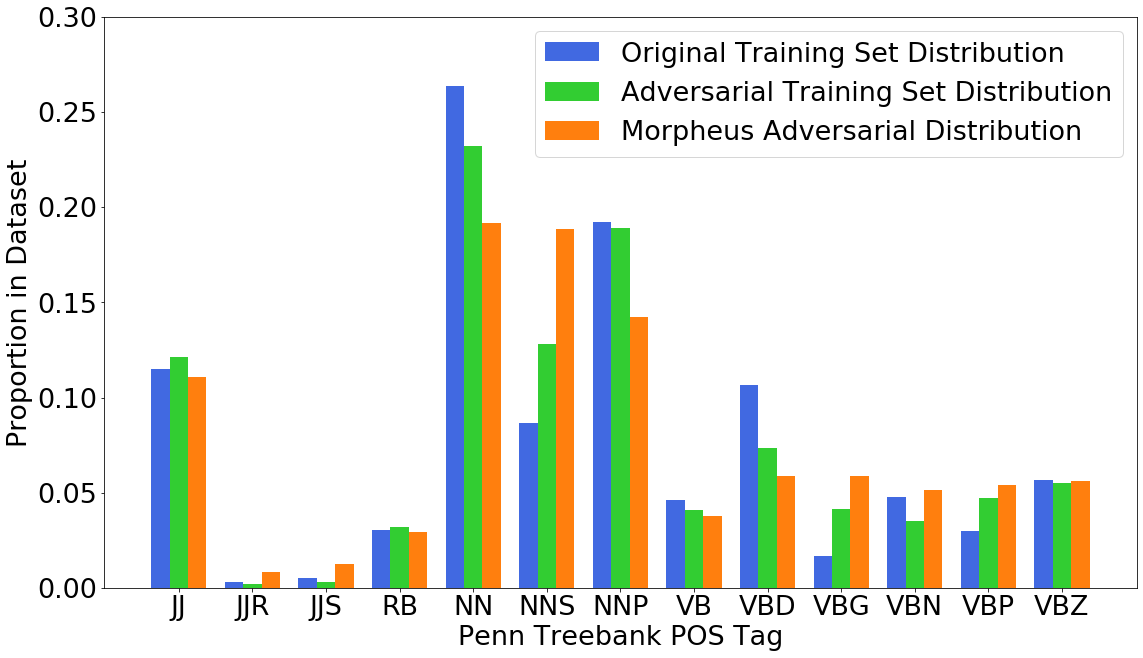}
        \caption{SQuAD 2.0 training set}
        \label{fig:training-dist}
    \end{subfigure}

    \caption{Comparison of inflectional distributions for SpanBERT$_{\text{SQuAD 2}}$. The adversarial distributions include only examples that degrade model performance. To make the best use of limited space, we omit the \texttt{RBR}, \texttt{RBS}, and \texttt{NNPS} tags since they do not vary much across distributions. Full figures in \Cref{appendix:misc_figs}.}
    \label{fig:orig_constr_unconstr}
\end{figure}
\begin{table*}[t!]
\small
    \centering
    \scalebox{0.95}{\begin{tabular}{c  c  c c c c c c c}
         \toprule
     \multicolumn{9}{c}{\textbf{SpanBERT$_{\text{SQuAD 2}}$ (F$_1$)}} \\
         \midrule
          & \multicolumn{2}{c}{Original} & & & & \multicolumn{3}{c}{Adversarially Fine-tuned} \\
          \cmidrule{2-3} \cmidrule{7-9}
          Dataset & Clean & \mor\ & & Epoch & & Clean & \mor$_{\text{orig}}$ & \mor$_{\text{adv}}$ \\
         \midrule
         \multirow{2}{*}{SQuAD 2.0 Ans} & \multirow{2}{*}{88.52} & \multirow{2}{*}{69.47 ($-$21.52\%)} & & 1 & & \textbf{86.80} & 85.17 ($-$1.87\%) & 82.76 ($-$4.65\%) \\
         & & & & 4 & & 86.15 & \textbf{84.93 ($-$1.41\%)} & \textbf{82.92 ($-$3.74\%)} \\
         \midrule
         \multirow{2}{*}{SQuAD 2.0 All} & \multirow{2}{*}{87.71} & \multirow{2}{*}{73.26 ($-$16.47\%)} & & 1 & &  86.00 & 84.72 ($-$1.48\%) & 82.41 ($-$4.17\%) \\
         & & & & 4 & &  \textbf{87.08} & \textbf{85.93 ($-$1.32\%)} & \textbf{84.71 ($-$2.72\%)} \\
         \midrule
         \multicolumn{9}{c}{\textbf{Transformer-big (BLEU)}} \\
         \midrule
         & \multicolumn{2}{c}{Original} & & & & \multicolumn{3}{c}{Adversarially Fine-tuned} \\
          \cmidrule{2-3} \cmidrule{7-9}
          Dataset & Clean & \mor\ & & Epoch & & Clean & \mor$_{\text{orig}}$ & \mor$_{\text{adv}}$ \\
         \midrule
         \multirow{2}{*}{newstest2014} & \multirow{2}{*}{43.16} & \multirow{2}{*}{20.57 ($-$56.25\%)} & & 1 & & 39.84  & \textbf{31.79 ($-$20.20\%)} & \textbf{31.43 ($-$21.10\%)} \\
         & & & & 4 & &  \textbf{40.60} & 31.99 ($-$21.20\%) & 30.82 ($-$24.08\%) \\
         \bottomrule
    \end{tabular}}
    \caption{Results from adversarially fine-tuning SpanBERT$_{\text{SQuAD 2}}$ and Transformer-big. \mor$_{\text{orig}}$ refers to the initial adversarial examples, while \mor$_{\text{adv}}$ refers to the new adversarial examples obtained by running \mor\ on the robust model. Relevant results from \Cref{tab:morpheus_results} reproduced here for ease of comparison.}
    \label{tab:adv_training_results}
\end{table*}

\section{Adversarial Fine-tuning}\label{sec:adv-train}

In this section, we extend the standard adversarial training paradigm \citep{goodfellow2015} to make the models robust to inflectional perturbations. Since directly running \mor\ on the entire training dataset to generate adversaries would be far too time-consuming, we use the findings from our experiments on the respective dev/test sets (\Cref{sec:experiments}) to create representative samples of good adversaries. This significantly improves robustness to inflectional perturbations while maintaining similar performance on the clean data.

We first present an analysis of the inflectional distributions before elaborating on our method for generating the adversarial training set.

\subsection{Distributional Analysis}

\Cref{fig:qa_constr_unconstr} illustrates the overall distributional differences in inflection occurrence between the original and adversarial examples found by \mor\ for SQuAD 2.0. Note that these distributions are computed based on the Penn Treebank (PTB) POS tags, which are finer-grained than the universal POS (UPOS) tags used to constrain \mor' search (\Cref{sec:experiments}). For example, a UPOS \texttt{VERB} may be actually be a PTB \texttt{VBD}, \texttt{VBZ}, \texttt{VBG}, etc.

We can see obvious differences between the global inflectional distributions of the original datasets and the adversaries found by \mor. The differences are particularly significant for the \texttt{NN}, \texttt{NNS}, and \texttt{VBG} categories. \texttt{NNS} and \texttt{VBG} also happen to be uncommon in the original distribution. Therefore, we conjecture that the models failed (\Cref{sec:experiments}) because \mor\ is able to find the contexts in the training data where these inflections are uncommon.

\subsection{Adversarial Training Set Generation}
Since there is an obvious distributional difference between the original and adversarial examples, we hypothesize that bringing the training set's inflectional distribution closer to that of the adversarial examples will improve the models' robustness.

To create the adversarial training set, we first isolate all the adversarial examples (from the dev/test set) that caused any decrease in F$_1$/BLEU score and count the number of times each inflection is used in this adversarial dataset, giving us the inflectional distribution in \Cref{fig:qa_constr_unconstr}.

Next, we randomly select an inflection for each eligible token in each \emph{training example}, weighting the selection with this inflectional distribution instead of a uniform one. To avoid introducing unnecessary noise into our training data, only inflections from the same UPOS as the original word are chosen. We do this 4 times per training example, resulting in an adversarial training set with a clean--adversarial ratio of $1:4$. 
This can be done in linear time and is \emph{highly scalable}. \Cref{alg:randominflect} in \Cref{app:adv_set_gen} details our approach and \Cref{fig:training-dist} depicts the training set's inflectional distribution before and after this procedure.

\paragraph{Fine-tuning vs. retraining.} Existing adversarial training approaches have shown that retraining the model on the augmented training set improves robustness \citep{belinkov2018synthetic,Eger2019TextPL,jin2019bert}. However, this requires substantial compute resources. We show that fine-tuning the pre-trained model for just a \emph{single} epoch is sufficient to achieve significant robustness to inflectional perturbations yet still maintain good performance on the clean evaluation set (\Cref{tab:adv_training_results}).

\subsection{Experiments}

\paragraph{SpanBERT.}
Following \citet{spanbert19}, we fine-tune SpanBERT$_{\text{SQuAD 2}}$ for another 4 epochs on our adversarial training set. \Cref{tab:adv_training_results} shows the effectiveness of our approach for SpanBERT$_{\text{SQuAD 2}}$. 

After just a single epoch of fine-tuning, SpanBERT$_{\text{SQuAD 2}}$ becomes robust to most of the initial adversarial examples with a $<2$-point drop in performance on the clean dev set. More importantly, running \mor\ on the robust model fails to significantly degrade its performance.

After 4 epochs, the performance on the clean SQuAD 2.0 dev set is almost equivalent to the original SpanBERT$_{\text{SQuAD 2}}$'s, however this comes at a slight cost: the performance on the answerable questions is slightly lower than before. In fact, if performance on answerable questions is paramount, our results show that fine-tuning on the adversarial training set for 1 epoch would be a better (and more cost effective) decision. Retraining SpanBERT adversarially did not result in better performance.

We also found that weighting the random sampling with the adversarial distribution helped to improve the robust model's performance on the answerable questions (refer to \Cref{tab:adv_qa_weighted_unweighted} in Appendix).

\paragraph{Transformer-big.}
Similarly, model robustness improves dramatically (56.25\% to 20.20\% decrease) after fine-tuning for 1 epoch on the adversarial training set with a \mytilde3 BLEU point drop in clean data performance (\Cref{tab:adv_training_results}). Fine-tuning for a further 3 epochs reduced the difference but made the model less robust to new adversarial examples.

We also experimented with using randomly sampled subsets but found that utilizing the entire original training set was necessary for preserving performance on the clean data (see \Cref{tab:subset_nmt} in Appendix).

\subsection{Discussion}

Our anonymous reviewers brought up the possibility of using grammatical error correction (GEC) systems as a defense against inflectional adversaries. Although we agree that adding a GEC model before the actual NLU/translation model would likely help, this would not only require an extra model---often another Transformer \citep{bryant-etal-2019-bea}---and its training data to be maintained, but would also double the resource usage of the combined system at inference time.

Consequently, institutions with limited resources may choose to sacrifice the experience of minority users rather than incur the extra maintenance costs. Adversarial fine-tuning only requires the NLU/translation model to be fine-tuned once and consumes no extra resources at inference time.

\section{Limitations and Future Work}

Although we have established our methods' effectiveness at both inducing model failure and robustifying said models, we believe they could be further improved by addressing the following limitations:

\begin{enumerate}[leftmargin=*]\itemsep0em
    \item \mor\ finds the distribution of examples that are adversarial for the target model, rather than that of \emph{real} L2 speaker errors, which produced some unrealistic adversarial examples.
    \item Our method of adversarial fine-tuning is analogous to curing the symptom rather than addressing the root cause since it would have to be performed for each domain-specific dataset the model is trained on.
\vspace{-0.2em}
\end{enumerate}

\noindent In future work, we intend to address these limitations by directly modeling the L2 and dialectal distributions and investigating the possibility of robustifying these models further upstream.

\section{Conclusion}
\label{sec:conclusion}

Ensuring that NLP technologies are inclusive, in the sense of working for users with diverse linguistic backgrounds (e.g., speakers of World Englishes such as AAVE, as well as L2 speakers), is especially important since natural language user interfaces are becoming increasingly ubiquitous.

We take a step in this direction by revealing the existence of linguistic bias in current English NLP models---e.g., BERT and Transformer---through the use of inflectional adversaries, before using adversarial fine-tuning to significantly reduce it. To find these adversarial examples, we propose \mor, which crafts plausible and semantically similar adversaries by perturbing an example's inflectional morphology in a constrained fashion, without needing access to the model's gradients. Next, we demonstrate the adversaries' effectiveness using QA and MT, two tasks with direct and wide-ranging applications, before validating their plausibility and semantic content with real humans.

Finally, we show that, instead of retraining the model, fine-tuning it on a representative adversarial training set for a single epoch is sufficient to achieve significant robustness to inflectional adversaries while preserving performance on the clean dataset. We also present a method of generating this adversarial training set in linear time by making use of the adversarial examples' inflectional distribution to perform weighted random sampling.

\section*{Acknowledgments}
We would like to express our gratitude to Lav Varshney, Jason Wu, Akhilesh Gotmare, and our anonymous reviewers for their insightful feedback on our paper, and friends who participated in our pilot studies. Samson is supported by Salesforce and the Singapore Economic Development Board under its Industrial Postgraduate Programme.

\bibliographystyle{acl_natbib}
\bibliography{ling,fairness,adversarial,misc,tasks}

\clearpage
\appendix
\label{sec:appendix}
\section{Examples of Inflectional Variation in English Dialects}
\label{appendix:dialect}
\paragraph{African American Vernacular English\\}
\citep{wolfram2004grammar}
\begin{itemize}
    \item They seen it.
    \item They run there yesterday.
    \item The folks was there.
\end{itemize}{}

\paragraph{Colloquial Singapore English (Singlish)\\} \citep{leimgruber2009singlish}
\begin{itemize}
    \item He want to see how we talk.
    \item It cover up everything in the floss. It’s not nice. It look very cheap.
    \item I want to shopping only.
\end{itemize}{}
\newpage
\section{More Details on Human Evaluation}
\begin{figure}[h]
    \centering
    \includegraphics[width=0.48\textwidth]{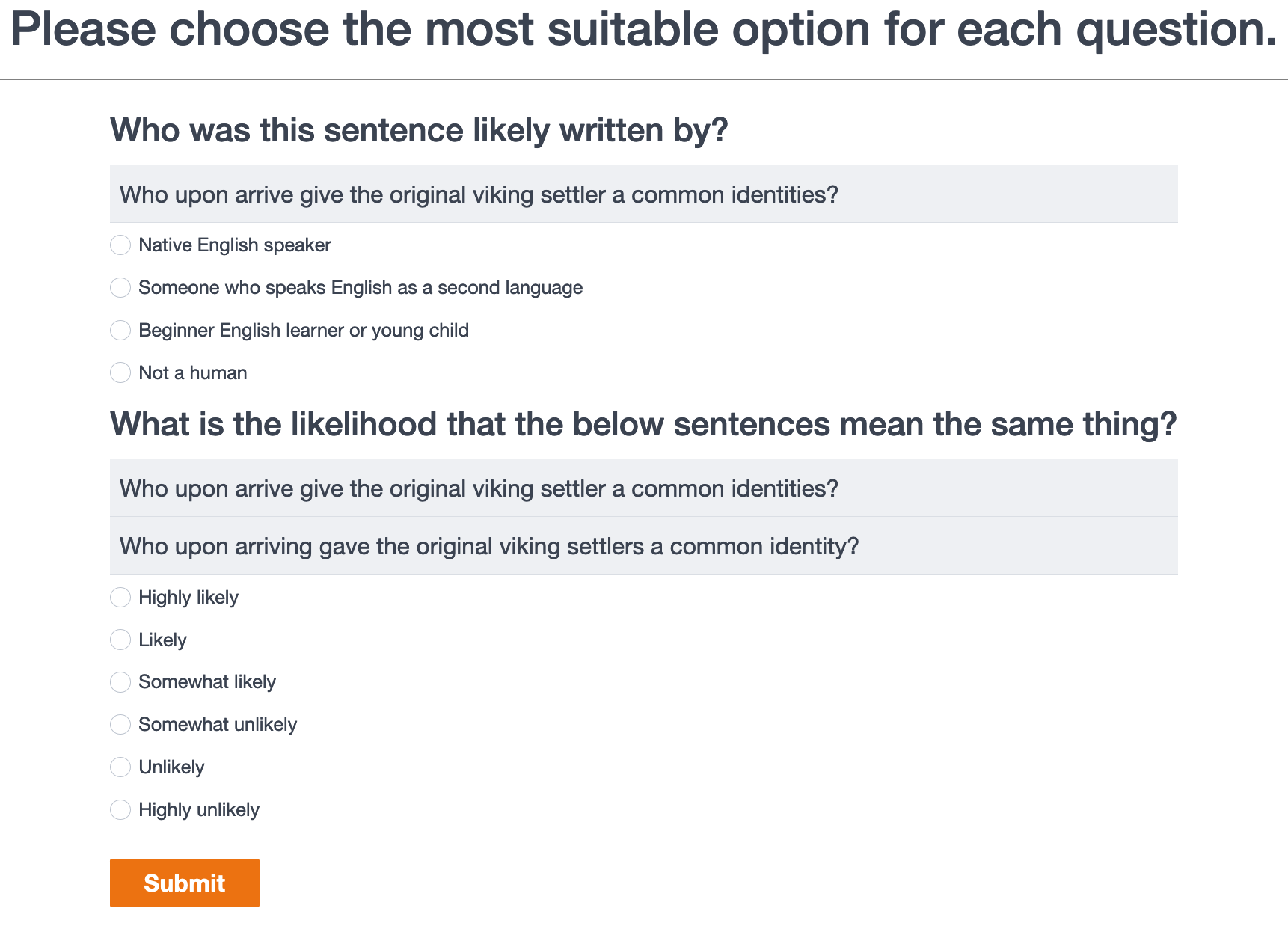}
    \caption{Amazon Mechanical Turk UI.}
    \label{fig:mturk_ui}
\end{figure}{}
\Cref{fig:mturk_ui} contains a screenshot of the UI we present to crowd workers. We intentionally prime Turkers by asking if the sentence could be written by an L2 speaker instead of directly asking for acceptability/naturalness ratings in order to ensure that they consider these possibilities.

We also do not use the Semantic Textual Similarity evaluation scheme \citep{agirre-etal-2013-sem}; during preliminary pilot studies, we discovered that annotators interpreted certain words in the scheme (e.g., ``information", ``details", and ``topics") considerably differently, introducing substantial noise into an already subjective judgement task.

\paragraph{Possible limitations.}It is possible that seeing the original sentence could affect the worker's judgment of the perturbed sentence's plausibility. However, we argue that this is not necessarily negative since seeing the original sentence would make it easier to spot perturbations that are just outright wrong (i.e., a human will not make that error regardless of their level of fluency).

\newpage
\section{Adversarial Training Set Generation}
\label{app:adv_set_gen}
\begin{algorithm}[h]
\small
\begin{algorithmic}
\Require Original instance $x$, hyperparameter $k$\\
Adversarial distribution $\gD_{adv}$
\Ensure Adversarial training dataset $X_x'$ for $x$
\State $X_x' \gets \{x\} $
\For {$i = 1$ to $k$}
\State $T \gets \Call{tokenize}{x}$

\ForAll {$i = 1, \ldots, |T|$}
    \If{$\Call{POS}{T_i}\in\! \{\text{\small NOUN},\text{\small VERB},\text{\small ADJ}\}$}
        \State $I \gets \Call{GetInflections}{T_i}$
        \State $T_i \gets \Call{RandomWeighted}{I, D_{adv}}$
    \EndIf
\EndFor

\State $x' \gets \Call{detokenize}{T}$
\State $X_x' \gets X_x' \cup \{x'\}$
\EndFor
\State \Return $X_x'$
\caption{RandomInflect}
\label{alg:randominflect}
\end{algorithmic}
\end{algorithm}

\newpage
\section{Tables and Figures}
\label{appendix:misc_figs}
\begin{table}[h]
\small
\centering
    \scalebox{0.87}{\begin{tabular}{c c c c}
        \toprule \multicolumn{4}{c}{\textbf{SpanBERT$_{\text{SQuAD 2}}$ (F$_1$)}} \\
         \midrule
         Dataset & Original & Morpheus$_{\text{seq}}$ & Morpheus$_{\text{parallel}}$ \\
         \midrule
         SQuAD 2.0 Ans & 88.52 & \textbf{69.47 (-21.52\%)} & 74.38 (-15.97\%) \\
         SQuAD 2.0 All & 87.71 & \textbf{73.26 (-16.47\%)} & 76.64 (-12.62\%) \\
         \midrule
         \multicolumn{4}{c}{\textbf{Transformer-big (BLEU)}} \\
         \midrule
         Dataset & Original & Morpheus$_{\text{seq}}$ & Morpheus$_{\text{parallel}}$ \\
         \midrule
         newstest2014 & 43.16 & \textbf{20.57 (-56.25\%)} & 20.85 (-51.69\%) \\
        
         \bottomrule
    \end{tabular}}
    \caption{Results of the parallel and sequential approaches to implementing \mor\ on SpanBERT$_{\text{SQuAD 2}}$ and Transformer-big.}
    \label{tab:parallel_seq}
\end{table}

\begin{table}[h]
    \centering
    \scalebox{0.85}{\begin{tabular}{c c c c}
         \toprule
     \multicolumn{4}{c}{\textbf{SpanBERT$_{\text{SQuAD 2}}$ (F$_1$)}} \\
         \midrule
         Weighted & Dataset & Clean & Morpheus$_{\text{orig}}$ \\
         \midrule
         \multirow{2}{*}{Yes} & SQuAD 2.0 Ans & 86.80 & 85.17 (-1.87\%) \\
         & SQuAD 2.0 All & 86.00 & 84.72 (-1.48\%) \\
         \midrule
         \multirow{2}{*}{No} & SQuAD 2.0 Ans & 84.52 & 83.15 (-1.62\%) \\
         & SQuAD 2.0 All & 87.12 & 86.03 (-1.25\%) \\
         \bottomrule
    \end{tabular}}
    \caption{Comparison of results from using weighted vs. uniform random sampling to the create adversarial training set for fine-tuning SpanBERT$_{\text{SQuAD 2}}$}
    \label{tab:adv_qa_weighted_unweighted}
\end{table}

\begin{table}[h]
    \centering
    \begin{tabular}{c c c c}
         \toprule
 \multicolumn{4}{c}{\textbf{Transformer-big (BLEU)}} \\
         \midrule
         Subset & Original & Clean & Morpheus$_{\text{orig}}$ \\
         \midrule
         $\frac{1}{20}$ & 43.16 & 30.90 & 24.95 \\
         \midrule
         $\frac{1}{4}$ & 43.16 & 36.59 & 29.46 \\
         \midrule
         Full & 43.16 & 40.60 & 31.99 \\
         \bottomrule
    \end{tabular}
    \caption{Results from adversarially fine-tuning Transformer-big on different subsets of the original training set.}
    \label{tab:subset_nmt}
\end{table}

\begin{figure*}
    \centering
    \includegraphics[width=\textwidth]{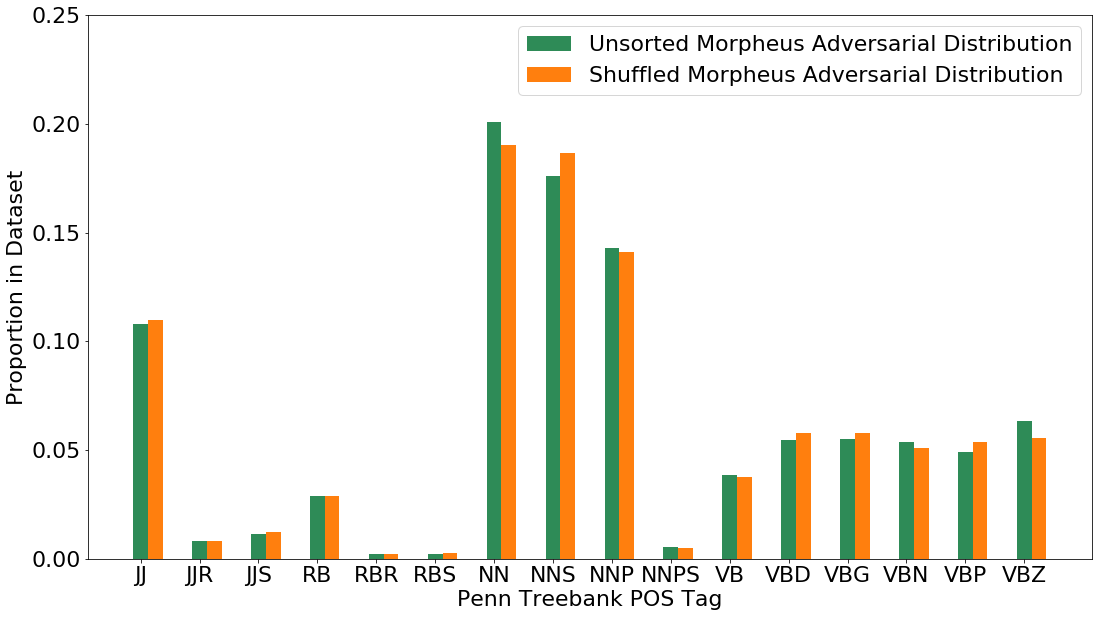}
    \caption{Effect of shuffling the inflection list on the adversarial distribution. We observe that shuffling the inflection list induces a more uniform inflectional distribution by reducing the higher frequency inflections and boosting the lower frequency ones.}
    \label{fig:shuffling}
\end{figure*}

\begin{figure*}
    \centering
    \begin{subfigure}{\textwidth}
        \includegraphics[width=\textwidth]{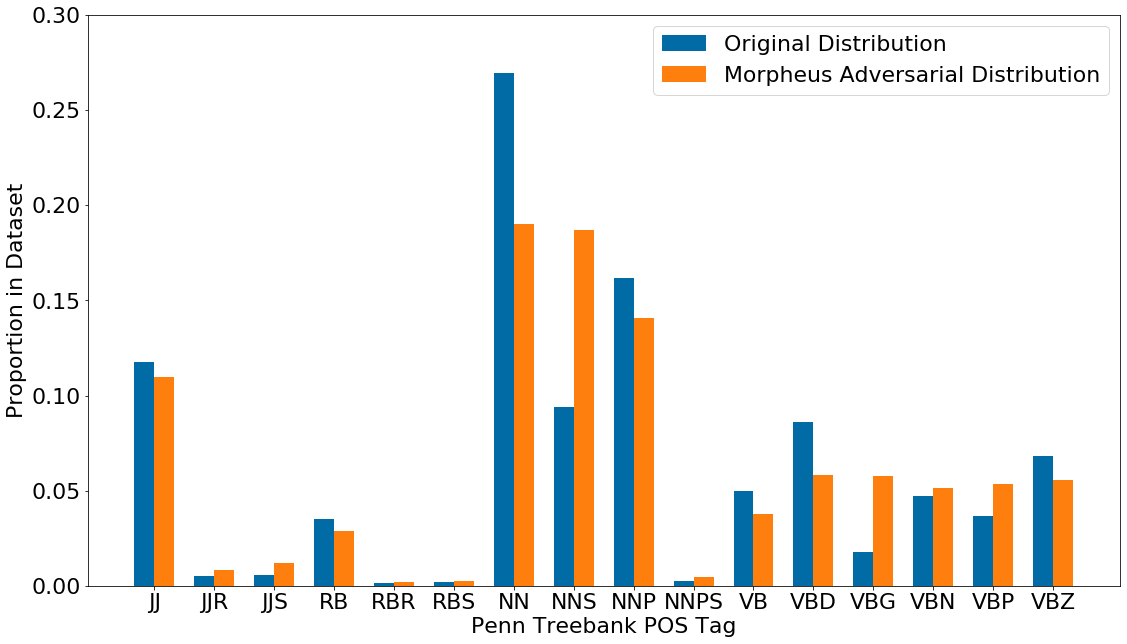}
        \caption{SQuAD 2.0 dev set}
    \end{subfigure}
     \begin{subfigure}{\textwidth}
        \includegraphics[width=\textwidth]{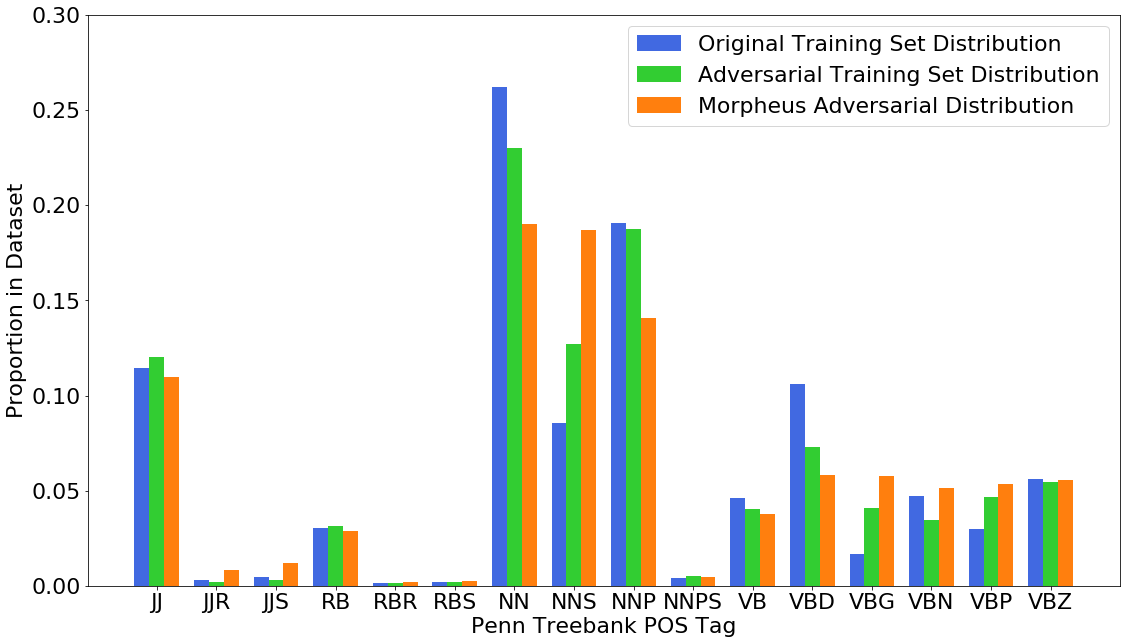}
        \caption{SQuAD 2.0 training set}
    \end{subfigure}
    \caption{Full versions of \Cref{fig:orig_constr_unconstr}}
\end{figure*}

\begin{table*}[h]
    \centering
    \begin{tabular}{l  p{0.75\textwidth}}
    \toprule
    Original Source & According to Detroit News, the queen of Soul will be performing at the Sound Board hall of MotorCity Casino Hotel on 21 December. \\
    Adversarial Source & \hl{Accorded} to Detroit News, the queen of Soul will be performing at the Sound Board hall of MotorCity Casino Hotel on 21 December. \\
    Original Translation & Selon Detroit News, la reine de Soul se produira au Sound Board Hall de l'hôtel MotorCity Casino le 21 décembre. \\
    \midrule
    Original Source & Intersex children pose ethical dilemma. \\
    Adversarial Source & Intersex child \hl{posing} ethical dilemma. \\
    Original Translation & Les enfants intersexuels posent un dilemme éthique. \\
    \midrule
    Original Source & The Guangzhou-based New Express made a rare public plea for the release of journalist Chen Yongzhou. \\
    Adversarial Source & The Guangzhou-based New \hl{Expresses} \hl{making} a rare public plea for the release of journalist Chen Yongzhou. \\
     Original Translation & Le New Express, basé à Guangzhou, a lancé un rare appel public en faveur de la libération du journaliste Chen Yongzhou. \\
     \midrule
     Original Source & Cue stories about passport controls at Berwick and a barbed wire border along Hadrian's Wall. \\
    Adversarial Source & Cue \hl{story} about passport controls at Berwick and a barbed \hl{wires} \hl{borders} along Hadrian's \hl{Walls}. \\
     Original Translation & Cue histoires sur le contrôle des passeports à Berwick et une frontière de barbelés le long du mur d'Hadrien. \\
    \bottomrule
    \end{tabular}
    \caption{Some of the adversaries that caused Transformer-big to output the source sentence instead of a translation.}
    \label{tab:eng_output}
\end{table*}

\end{document}